\pdfoutput=1

\documentclass[11pt]{article}

 \usepackage[]{acl}

\usepackage{times}
\usepackage{graphicx}
\usepackage{latexsym}
\usepackage{svg}
\usepackage{color,colortbl} 
\usepackage{booktabs}
\usepackage{soul}
\usepackage{multirow}
\usepackage{tikz}
\usetikzlibrary{automata,arrows}
\usepackage{tikzscale}          
\usepackage{pgfplots}
\usepackage{pgfplotstable}
\pgfplotsset{compat=1.17}
\usepackage{enumitem}
\usetikzlibrary{
  matrix,
}
\tikzset{
  every edge/.append style = {thick},
  mynode/.style={minimum size=0.7cm,shape=circle,draw=black,thick}
}
\usepackage{array}
\newcolumntype{L}[1]{>{\raggedright\let\newline\\\arraybackslash\hspace{0pt}}m{#1}}
\newcolumntype{C}[1]{>{\centering\let\newline\\\arraybackslash\hspace{0pt}}m{#1}}
\newcolumntype{R}[1]{>{\raggedleft\let\newline\\\arraybackslash\hspace{0pt}}m{#1}}

\colorlet{soulcyan}{cyan!30}
\DeclareRobustCommand{\hlcyan}[1]{{\sethlcolor{soulcyan}\hl{#1}}}
\colorlet{soullime}{lime!30}
\DeclareRobustCommand{\hllime}[1]{{\sethlcolor{soullime}\hl{#1}}}
\colorlet{soulred}{red!30}
\DeclareRobustCommand{\hlred}[1]{{\sethlcolor{soulred}\hl{#1}}}
\colorlet{soulorange}{orange!30}
\DeclareRobustCommand{\hlorange}[1]{{\sethlcolor{soulorange}\hl{#1}}}
\colorlet{soulpurple}{purple!30}
\DeclareRobustCommand{\hlpurple}[1]{{\sethlcolor{soulpurple}\hl{#1}}}
\usepackage[T1]{fontenc}

\usepackage[utf8]{inputenc}

\usepackage{microtype}

\definecolor{intcolor}{HTML}{848FA2}
\definecolor{english}{HTML}{CC2D35}
\definecolor{hindi}{HTML}{058ED9}

\newcommand{\hindi}[1]{{\color{hindi}#1}}
\newcommand{\english}[1]{{\color{english}#1}}
\newcommand{\error}[1]{{\color{red}#1}}
\newcommand{\name}{CST5}
\newcommand{\datasetname}{Hinglish-TOP }
\title{CST5: Data Augmentation for Code-Switched Semantic Parsing}

\author{
    Anmol Agarwal\thanks{This work was done during an internship at Google}, Jigar Gupta, Rahul Goel, Shyam Upadhyay \\ \textbf{Pankaj Joshi}, \textbf{Rengarajan Aravamudhan}
    \\
    Google Assistant\\

    \texttt{\{okanmol,jigargupta,goelrahul,shyamupa,pnkaj,rengarajan\}@google.com}
}

\begin{document}
\maketitle
\begin{abstract}
Extending semantic parsers to code-switched input has been a challenging problem, primarily due to a lack of supervised training data.  In this work, we introduce CST5, a new data augmentation technique that finetunes a T5 model using a small seed set ($\approx$100 utterances) to generate code-switched utterances from English utterances. We show that CST5 generates high quality code-switched data, both intrinsically (per human evaluation) and extrinsically by comparing baseline models which are trained without data augmentation to models which are trained with augmented data. 
Empirically we observe that using CST5, one can achieve the same semantic parsing performance by using up to 20x less labeled data. 
To aid further research in this area,  we are also releasing (a) Hinglish-TOP, the largest  human annotated code-switched semantic parsing dataset to date, containing 10k human annotated Hindi-English (Hinglish) code-switched utterances, and (b) Over 170K CST5 generated code-switched utterances from the TOPv2 dataset. 
Human evaluation shows that both the human annotated data as well as the CST5 generated data is of good quality.  
\end{abstract}


\section{Introduction}

Code-switching (CS) occurs when a speaker alternates between two or more languages within a single conversation. 
With the increasing ubiquity of voice-based assistants (e.g., Google Assistant, Alexa), the ability to parse code-switched utterances has become one of the key challenges towards building multilingual conversational agents. 



Unfortunately, majority of semantic parsing datasets are in English, with very few datasets exhibiting code-switching.
While data collection efforts~\cite{einolghozati2021volumen,mehnaz2021gupshup} have been made to bridge this gap, collecting such datasets requires time-consuming and expensive human annotations from raters who are proficient in multiple languages, making it difficult to obtain large datasets. 
Another line of work generates artificial code-switched data, either using parallel sentences in the two languages as supervision~\cite{winata2019} or learning a generative language model from large code-switched corpora~\cite{chang2019codeswitching}. However, these approaches assume the availability of large corpora (>50k sentences), either parallel, monolingual or code-switched in the languages of interest, which may not be available for the domains of interest.  
\begin{table}
    \footnotesize
    \centering
    \begin{tabular}{L{3cm}L{4cm}}
         \toprule
         \rowcolor[gray]{0.97} Original English Sentence & {\name~ Generated Code Switched Utterance} \\
         \midrule
          {Create a new alarm for 9AM on monday 18th June} & \english{Monday 18th June} \hindi{ko subah 9 bajhe ke liye ek naye} \english{alarm} \hindi{ko} \english{create} \hindi{karen} \\
         {Remove Jim from my reminder to party next wednesday} & \hindi{Agle} \english{wednesday} \hindi{ko} \english{party} \hindi{ke liye} \english{Jim} \hindi{ko mere} \english{reminder} \hindi{se hata den} \\
         \bottomrule
    \end{tabular}
    \caption{A few examples of code switched data generated by CST5. Even with 100 seed examples, the model generates good quality data across multiple domains.  \english{English} and \hindi{Hindi} tokens are color coded. }
    \label{examples}
\end{table}
\begin{figure*}
  \centering
  \includegraphics[width=\textwidth,height=8cm]{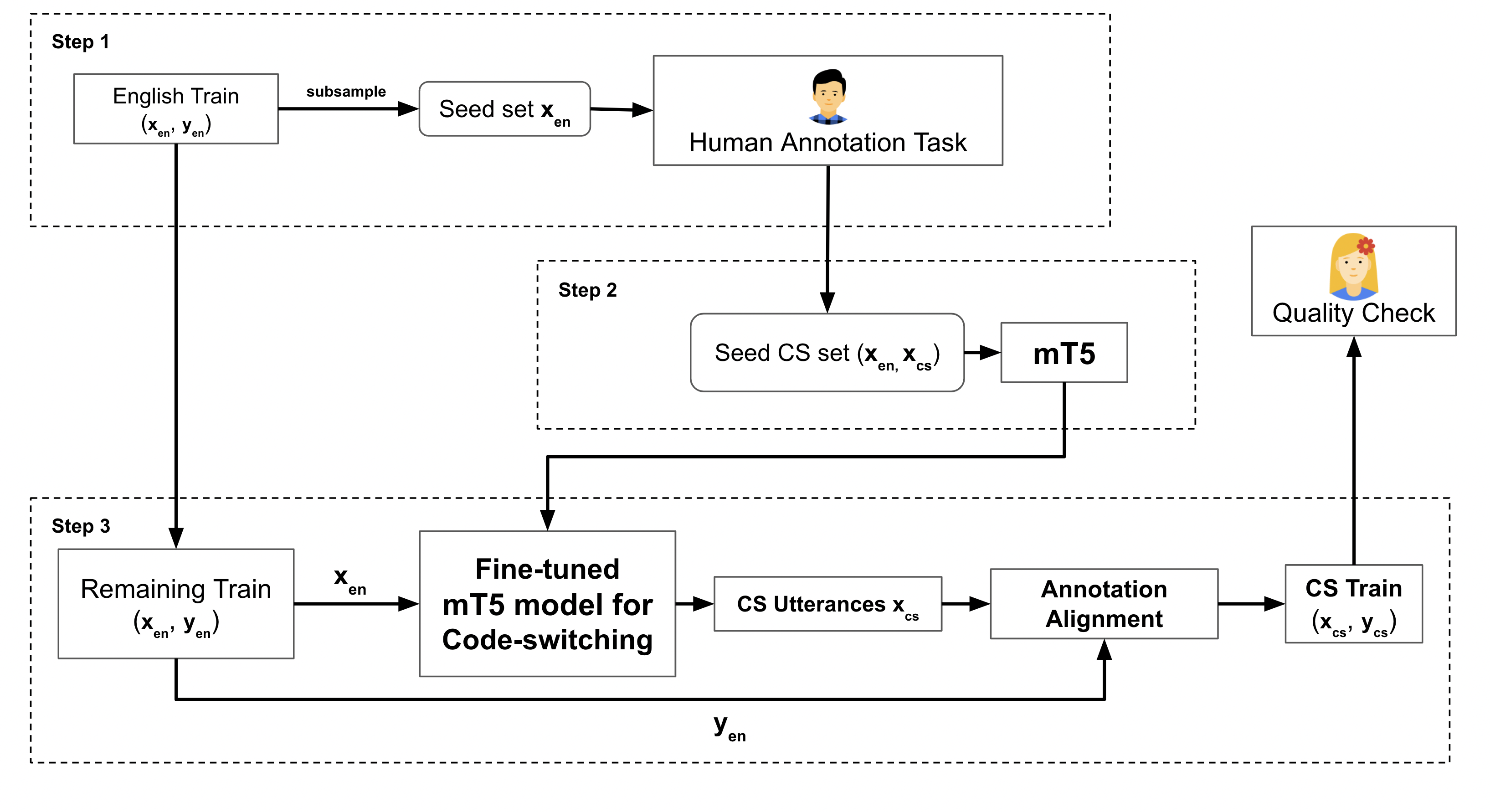}
  \caption{An overview of our approach. We first use human annotators to annotate utterances. We use these to fine tune a mT5 model for code switching and then use the rest of the monolingual corpus to generate more code switched data. }
  \label{block_1}
\end{figure*}

\begin{figure}
  \centering
  \includegraphics[width=\linewidth]{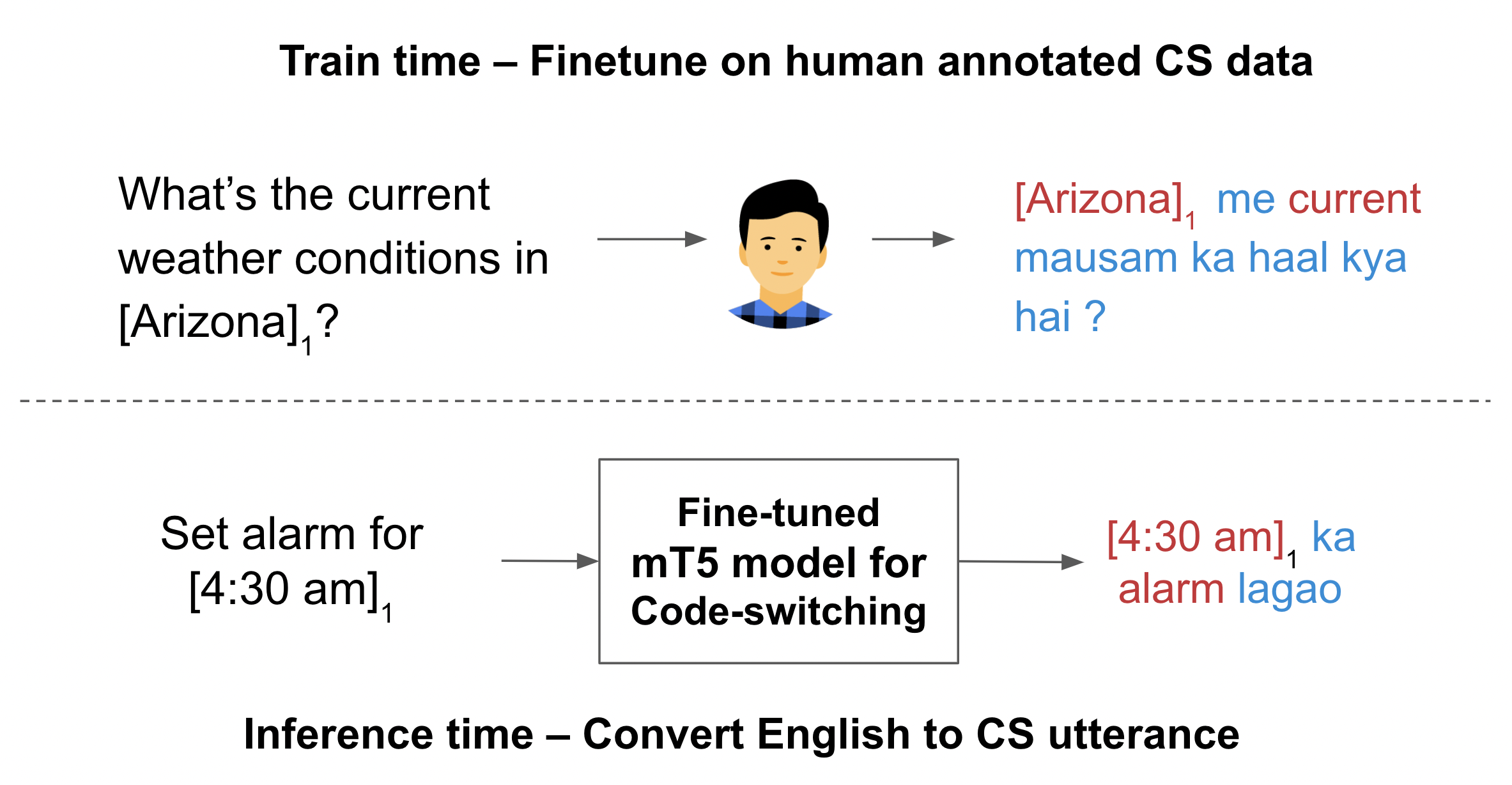}
  \caption{We finetune mT5 on parallel human-annotated data and do inference using the finetuned mt5 model using the monolingual data}
  \label{block_2}
\end{figure}

In this paper we propose \name, a strategy to obtain code-switched labeled data by generating code switched utterances from a English semantic parsing dataset, using only a small ($\approx$100 examples) seed set of (English, Code-switched) example pairs to learn the transformation. 
This allows us to effectively leverage existing semantic parsing datasets in English to derive supervision for code-switched semantic parsing. Using \name, we achieve an average 25\% improvement on the exact match~(EM) accuracy, equivalent to having a order of magnitude more training data. A few examples of the CS utterances produced by CST5 are shown in Table~\ref{examples}.   

Our work is inspired by the growing popularity of using large pre-trained language models (LM) (such as BERT~\cite{devlin2018bert}, T5~\cite{raffel2019exploring} etc.) as a means of data augmentation~\cite[inter alia]{kumar2019,AnabyTavor2019,lee2021neural}.
Particularly, \name~ relies on mT5~\cite{xue2020mt5}, a large pre-trained multilingual LM, which is fine-tuned to perform code-switching using the small seed set, and then applied on a large monolingual semantic parsing dataset in English to generate code switched utterances. 
By mapping the labeled semantic parse for the English utterance, one can recover the semantic parse to the code-switched utterance, thus obtaining a labeled code-switched semantic parsing dataset. 
We observe that the mT5 model is well suited to the task of code-switching and the generated code-switched utterances were deemed of high quality upon manual inspection by human raters, judged to be over 89\% semantically equivalent to the original English utterance, and 98\% natural. 

We use CST5 to generate synthetic code switched data and measure the effect of data augmentation on parser performance with varying seed sizes and across various semantic parsing domains.
We observe a 20x data reduction for code switching when using the XXL model. Even with a 100 seed examples, our approach matches the performance of a model trained with 2000 examples for the XXL mT5 parser. To summarize, in this work: 
\begin{itemize}[noitemsep,topsep=0pt,parsep=0pt,partopsep=0pt]
    \item We introduce a novel data augmentation technique to generate synthetic code switched data for semantic parsing. Using a small seed set of parallel data, the technique converts mono-lingual data to code switched data. 
    \item  We did additional human evaluation of the generated data showing high data quality.
    \item  We conducted experiments on TOPv2 as well as CSTOP\cite{einolghozati2021volumen} showing significant gains in EM accuracy.
    \item  We release a code-switched task-oriented semantic parsing dataset, containing over 10k Hindi + English human annotated CS utterances (Hinglish-TOP) along with over 170K synthetically generated CS utterances and their corresponding semantic parse. 
\end{itemize}


\section{Related Work}
Our work is at the intersection of two areas in NLP:  data augmentation for code switching and conversational semantic parsing.

\subsection{Data Augmentation for Code Switching}
Data augmentation is seen as a flexible and model independent tool for performance improvement by synthetically generating training data for the task of interest \citep[\textit{inter alia}]{jia2016data,andreas2019good,akyurek2020learning}.
We discuss some approaches for synthetically generating code switched data below.

\citet{winata2019} generated code switched utterances using parallel sentences in the two languages as supervision.
\citet{chang2019codeswitching} proposed an unsupervised method to generate code-switching sentences from monolingual sentences using GANs. Linguistically motivated data augmentation has also been explored in \citet{lee2019linguistically} and \citet{pratapa2018language}. Such techniques require expert knowledge or large amounts of parallel data which might be hard to get in a new domain. Instead our work uses small seed sets to generate synthetic data which is more scalable.

\paragraph {Language Model based Data Augmentation} \citet{kumar2021data} used pre-trained transformer models, like BART~\cite{lewis-etal-2020-bart} and GPT-\cite{radford2019language}, for general data augmentation. Our work  leverages  large pretrained language models (LMs)~\cite{raffel2019exploring,brown2020language} to generate new synthetic examples. Our approach draws inspiration from recent models like ex2~\cite{lee2021neural}, mT5~\cite{xue2020mt5} and is most similar to the translate and fill approach suggested by ~\citet{nicosia2021translate}. In our work instead of fine tuning the LM to generate translations, we fine tune it to generate CS utterances.

\subsection{Conversational Semantic Parsing} 

Today most commercial~\cite{lialin2020update} conversational semantic parsing systems utilize hierarchical representations, such as the TOP representation ~\cite{gupta2018semantic}, which is typically modelled using sequence to sequence task ~\cite{rongali2020don,shaw2020compositional}. Our semantic parsing model is most similar to ~\citet{cole2021graph} where we use a T5 model as the base semantic parser.

\paragraph{Code-Switched Semantic Parsing} Traditionally, CS has been explored in word-level language identification~\cite{molina2019overview} and named entity recognition~\cite{aguilar2019named} and shallow parsing~\cite{sharma2016shallow,bhat2018universal}.
Recent works like ~\citet{duong2017multilingual} introduced a CS test set for semantic parsing,  ~\citet{Samanta2019ADG} introduce a variational auto-encoder based generation technique, some other works take inspiration from machine translation~\cite{tarunesh2021} and adversarial networks~\cite{Chandu2020}. These works rely on  specialized architectures and only produce code switched sentences, whereas our work simply relies on a LM finetuning and outputs both sentences and semantic parses. 
 
More recently, ~\citet{einolghozati2021volumen} released a Spanglish semantic parsing dataset named CSTOP and ~\citet{mehnaz2021gupshup} released a Hinglish conversation summarization dataset named GupShup. Similar to the previous works, we release a TOPv2 derived code-switched dataset TOP-Hinglish. We also release the synthetic augmented dataset which is an order of magnitude larger than any such previous dataset. 


\section{The Code Switched Dataset }

\subsection{Collecting Hinglish-TOP}
\begin{table*}
\footnotesize
\centering
\begin{tabular}{lL{6cm}L{6cm}}
\textbf{Domain} & \textbf{Pre-processed English Utterance (Input)} & \textbf{Code-Switched Utterance (Output)}\\
\toprule
\scriptsize{Navigation (14.5\%)} & {What's the traffic like on [Long Island]$_1$ going to [the Hamptons]$_2$ [tonight]$_3$} & \hindi{[Aaj raat]$_3$} \english{[Hamptons]$_2$} \hindi{jaate hue} \english{[Long Island]$_1$} \hindi{par} \english{traffic} \hindi{kaisa hoga} \\
\scriptsize{Weather (16.6\%)} & {Whats the current weather conditions in  [Arizona]$_2$ ?} & \english{[Arizona]$_2$} \hindi{me} \english{current} \hindi{mausam ka haal kya hai ?} \\
\scriptsize{Reminder (14.1\%)} & {Remind [me]$_1$ to [send out the company meeting notes]$_2$ [at 5 pm today]$_3$.} & \hindi{[Muje]$_1$ aaj [shaam 5 baje]$_3$} \english{[company meeting note} \hindi{ko} \english{send]$_2$} \hindi{karne ka} \english{remind} \hindi{karaye} \\
\scriptsize{Alarm (18.2\%)} & {Set me an alarm [every Thursday at 5AM]$_1$ [until the 1st July]$_2$} & \hindi{Muje [}\english{1 july} \hindi{tak]$_2$ ke liye [har} \english{thursday} \hindi{ko subah 5 baje]$_1$ ka} \english{alarm} \hindi{set kare} \\
\scriptsize{Music (10.8\%)} & {Can I listen to the 5th [song]$_1$ on the [album]$_2$ one more time ?} & \hindi{kya mai} \english{[album]$_2$} \hindi{par} \english{5th [song]$_1$} \hindi{ek bar fir sun sakta hoon ?} \\
\scriptsize{{Timer} (10.9\%)} & {I want you to change the [timer]$_1$ to a [30 minutes]$_2$ instead of [15 minutes]$_3$.} & \hindi{Mai chahta hu ki aap} \english{[timer]$_1$} \hindi{ko} \english{[15 minutes]$_3$} \hindi{ke bajaye} \english{[30 minutes]$_2$} \hindi{me badal den.} \\
\scriptsize{Event (6.9\%)} & {When does [cider tasting]$_1$ season start in [ Woodinville, Washington]$_2$} & \english{[Woodinville , Washington]$_1$} \hindi{me} \english{[cider tasting]$_1$ season} \hindi{kab shuru hoga} \\
\scriptsize{Messaging (8.0\%)} & {Send a message to [Diana]$_1$ and [Rich]$_2$ that [I will be earlier than expected to the library]$_3$.} & {\english{[Diana]$_1$} \hindi{aur} \english{[Rich]$_2$} \hindi{ko} \english{message} \hindi{bhejo ki [mai} \english{library time} \hindi{se pehle pahunch jaunga]$_3$}} \\
\bottomrule
\end{tabular}
\caption{\label{annotator_examples}
Examples of CS utterances obtained from annotators for the 8 domains in the TOPv2 dataset. The relative sizes of each domain is in parenthesis. The CS utterance is aligned with the English utterance's parse to identify the CS utterance's semantic parse (Section~\ref{annot_sec}).}
\end{table*}
For the Hinglish-TOP data collection, we randomly sampled a set of 10,896 utterances from the TOPv2 dataset ~\cite{chen2020low}. These utterances are distributed across 8 domains (as present in the original TOPv2 dataset). We used a skewed train, validation and test set distribution to focus on getting a bigger test set. Following this we sampled around 2993, 1390 and 6513 utterances from the train, validation and test splits of the original TOPv2 dataset. We observed that there were overlaps across the three splits in the original TOPv2 dataset which we also preserved as is. A domain-wise breakdown of the annotated CS data and a few examples can be found in Table~\ref{annotator_examples}.

Our data collection process can be broken down into the steps described below.  For these steps we preserved the domain and split information for the sampled  utterances. Our approach is also detailed in Figure~\ref{block_1}.

\subsubsection{Pre-Processing}

As a pre-processing step, the slots and intents present in the semantic parse of the English utterance were marked within the utterance using special span identifiers (span-ID), which aligns their position in the semantic tree. By marking these spans we know what words in the sentence correspond to the leaf arguments and ask the raters to mark the same in the code-switched query. Some examples of pre-processed English utterances can be seen in Table \ref{annotator_examples}, 2$^{nd}$ column. The pre-processing step allows us to reconstruct the semantic parse for the code-switched utterance, described next. 

\subsubsection{Code-switching and Alignment} 
\label{annot_sec}
Native Hindi-English speakers were chosen as human annotators for code-switching the English utterances. We asked the the annotators to generate a hinglish code switched utterance for the given sentence  while preserving the naturalness and semantic equivalence of the code-switched utterance. This task was done by 3 annotators.  A primary annotator whose task is to code-switch the English utterance.  Additionally, we had one more annotator who checked the utterances for naturalness and semantic equivalence. We keep the utterance when both the checker agreed on the naturalness and semantic equivalence. The definition of naturalness and semantic equivalence is same as in Section~\ref{sec:data_quality}.   Some examples of code-switched utterances after the data annotation task are shown in Table \ref{annotator_examples}. This step is shown as Step 1 in Figure \ref{block_1}. 

\paragraph{Annotation Alignment} After obtaining human annotated code-switched utterances with the slot spans ($3^{rd}$ column Table~\ref{annotator_examples}), we transfer the slot names (e.g., $date\_time$, $alarm\_name$ etc.) from the English parse to the code-switched utterance to obtain human labeled data for code-switched semantic parsing. As we preserved the slot spans (Table \ref{annotator_examples}) as the part of the parsing task there is a 1:1 mapping between the english parse the code switched utterance. The example below denotes this step, we map each date\_time slot to a unique span id to get a 1:1 alignment. 

\begin{tabular}{C{7.2cm}}
\footnotesize
[ Set alarm [ for 4:30 am on Tuesday ]$_{date\_time}$ 
and [ Thursday ]$_{date\_time}$ of next week ]$_{create\_alarm}$
     \\
\footnotesize {Set alarm [ for 4:30 am on Tuesday ]$_1$ and [ Thursday ]$_2$ of next week}
\end{tabular}

At the end of this step, we had collected 10k CS utterances, along with the semantic parses derived using the alignment above. Details on our human annotated \datasetname~dataset are in Table~\ref{dataset_stats}.

\begin{table*}
\footnotesize
\centering
\begin{tabular}{L{4cm}L{5cm}L{6cm}}
\toprule
\textbf{English Utterance} & \textbf{Code-switched Utterance} & \textbf{Rating}  \\
\midrule
{What is the weather in Canada ?} & \english{Canada} \hindi{me mausam kaisa hai ?} & \textbf{Natural \& Semantically similar}  \\
\midrule
{How long is left before my alarm goes off ?} & \hindi{Mera} \english{alarm} \hindi{kab tak band rahega ?} & \textbf{Natural but Not Semantically similar}: English utterance is asking for time remaining, while CS utterance is asking for trigger time  \\
\midrule
{Remind me the first Tuesday of every month that bulk trash pick up is the next morning .} & \hindi{Muje} \english{every month first Tuesday} \hindi{ko} \english{remind} \hindi{karaye ki} \english{bulk trash pick up is} \hindi{agley subha hai .} & \textbf{Not Natural but Semantically similar}: Phrase "bulk trash  pick up is agley subha hai" is un-natural\\
\midrule
{How long of a delay is traffic causing} & \english{traffic} \hindi{kaisa} \english{delay causing} \hindi{hai} & \textbf{Not Natural \& Not Semantically similar}  \\
\bottomrule
\end{tabular}
\caption{\label{natural_similarity_examples}
Examples of CS utterances generated by the XXL mT5 model, along with their human ratings.}
\end{table*}

\begin{table}
\centering
\footnotesize
    \begin{tabular}{lr}
    \toprule
        English Vocabulary size &  4857\\
    \midrule
        Romanized Hindi Vocabulary size & 1931\\
    \midrule
        Total utterances & 10,896\\
    \midrule
        Avg. \# of Hindi tokens per utterance & 4.36\\
    \midrule
        Avg. \# of English tokens per utterance & 3.82\\
    \midrule
        Avg. \# of CS points per utterance & 3.56\\
    \bottomrule
    \end{tabular}
    \caption{Dataset statistics for the human annotated \datasetname~ dataset. }
    \label{dataset_stats}
\end{table}
\subsection{CS Data Generation using mT5} \label{augment_sec}

\begin{table*}
\footnotesize
\centering
\begin{tabular}{L{4.5cm}L{5cm}L{5cm}}
\toprule
\textbf{English Utterance} & \textbf{Generated Code-Switched Utterance} & \textbf{Error Type}\\
\midrule
{[ 9 pm ]$_1$ [ appointment for photos ]$_2$ and remind [ me ]$_3$ [ an hour before ]$_4$ } & {[ mujhe ]\error{$_3$} [ 9 pm ]$_1$ ko [ photos ke liye appointment ]$_2$ hai aur [ mujhe ]\error{$_3$} [ ek ghante pehle ]$_4$ yaad dilaayen} & {Unequal number of argument spans in English (one [ me ]$_3$) and code-switched query (two [ mujhe ]$_3$)} \\
\midrule
{play [ song ]$_1$ [ Heart is on fire ]$_2$ on [ spotify ]$_3$} & {[ spotify ]$_3$ par [ song ]$_1$ [ Heart is on fire ]\error{$_{two}$} ko bajao} & {Improper span ID formatting for Code-Switched utterances} \\
\midrule
{Change [ banking ]$_1$ reminders [ from ]$_2$ [ once a week ]$_3$ [ to ]$_4$ [ twice a week ]$_5$} & {[ banking ]$_1$ reminders ko \error{[} \error{[} ek bar har week \error{]$_3$} [ dohrayen ]$_4$} & {Unequal number of opening and closing spans for Code-Switched utterances} \\
\midrule
{Remind [ me ]$_1$ to [ email ]$_2$ [ Michelle ]$_3$ [ on Tuesday ]$_4$ [ about ]$_5$ [ the recital ]$_6$} & {[ Mujhe ]$_1$ [ Tuesday ko ]\error{$_7$} [ Michelle ]$_3$ ko [ email ]$_2$ karne ke liye yaad dilaayen} & {Non-matching span IDs in english and code-switched utterances} \\
\bottomrule
\end{tabular}
\caption{\label{error}
Examples of errors that were automatically filtered from the mT5 generated code switched utterances. }
\end{table*}

\begin{table}
\footnotesize
\centering
\begin{tabular}{lcc}
\toprule
\textbf{Data source} & \textbf{Naturalness} & \textbf{Semantic Equiv.} \\
\midrule
{Human} &  {99.2\% (0.72)} & {97.7\% (0.69)} \\
\midrule
{mT5 Small} &  {86.1\% (0.8)} & {43.7\% (0.88)} \\
{mT5 XXL} &  {98.1\% (0.95)} & {89.5\% (0.73)} \\
\bottomrule
\end{tabular}
\caption{\label{data_quality}
Naturalness and Semantic Equivalence Statistics from Data Quality Task. The $\kappa$ values for rater agreement is listed in brackets.}
\end{table}

In order to generate synthetic code-switched utterances, we fine-tuned a mT5 ~\cite{xue2020mt5} model for the task of code-switching the English utterances using the training data obtained in the previous step. The input to the T5 model is pre-processed English utterance (second column Table~\ref{annotator_examples}) and the output is the code-switched utterance containing slot spans (3$^{rd}$ column Table~\ref{annotator_examples}).  The TOPv2 dataset originally had over 180K utterances, of which we manually annotated only 10.8K (3K train, 1.3K validation and 6.5K test split) utterances. The annotated train set was used to fine-tune the model. The remaining  utterances  from TOPv2  was used as monolingual input to the mT5 model to  generate synthetic code-switched utterances.

Since mT5 is a text to text model, the generated synthetic code-switched utterances were not always structurally correct. This  resulted in errors when trying to align the output with the English semantic parse as mentioned in Section~\ref{annot_sec}. 

\subsubsection{Data Filtering}
\begin{table}[ht]
\centering
\footnotesize
\begin{tabular}{L{2cm}C{2cm}C{2cm}}
\toprule
\textbf{Seed Set Size} & \textbf{Small mT5 Throughput} & \textbf{XXL mT5 Throughput} \\
\midrule
100 & 47.3\%	& 82.0\% \\
500 & 64.4\% & 93.7\% \\
1k & 72.8\% & 96.1\% \\
2k & 89.7\% & 97.5\% \\
3k & 92.7\% & 98.3\% \\
\bottomrule
\end{tabular}
\caption{Percentage of input English queries that get code switched  post data-filtering with varying number of seed examples for both XXL and small model. We use the remaining TOPv2 dataset (173,042 utterances) for generating the CS utterances.}
\label{synthetic_data_stat}
\end{table}
Sometimes the code-switched utterances generated from the fine-tuned mt5 model have structural errors which prevent aligning the semantic parse for these code-switched utterances to their English counterparts. We filter out such utterances using a syntactic rule based filter. In particular we removed examples which contain:

\begin{itemize}[noitemsep,topsep=0pt,parsep=0pt,partopsep=0pt]
\item  Unequal number of argument spans in English and code-switched utterances.
\item  Improper span-ID formatting (when the span-ID could not be extracted) for code-switched utterances.
\item  Unequal number of opening and closing spans for code-switched utterances.
\item  Non-matching span-ID in English and code-switched utterances.
\end{itemize}
Examples of these errors can be seen in Table~\ref{error}. 

Finally, post data-filtering, we align the semantic parse annotations of the English utterance to the code-switched utterances using the same approach as in Section \ref{annot_sec}. 
The amount of syntactically correct synthetically generated data different seed set sizes is shown in Table~\ref{synthetic_data_stat}. Post data-filtering using 100 fine-tuning examples, the small mT5 model had a yield of 47.3\% on the remaining TOPv2 dataset, while the XXL model successfully converted 82.0\% of the English utterances.

\subsection{Data Quality Assessment} \label{sec:data_quality}
We conduct both intrinsic and extrinsic evaluation of the generated synthetic data. The results of extrinsic evaluation are presented in Section~\ref{results}. In this section we focus on 2 main questions:

\textbf{\textit{What is the data quality of the human annotated data?}}
To measure the quality of the CS data we created a {second} annotation task in which speakers proficient in both English and Hindi rated the CS utterances for their \textit{naturalness} and \textit{semantic equivalence} to the original English utterance. 
For naturalness, raters were just shown the CS utterance and were asked to answer the question: \textit{Would a native speaker utter this utterance naturally in a conversation?} For semantic equivalence, raters were shown both the English and the CS utterance and were asked: \textit{Do these sentences convey exactly the same meaning?} This serves as an upper bound on naturalness and semantic equivalence. Example ratings for naturalness and semantic equivalence are shown in Table~\ref{natural_similarity_examples}.

We did this study on a sample of 2021 utterances. Each CS utterance was rated by two human raters, and conflicts in opinion were resolved by using a third rater and taking the majority opinion.  For the human annotated data (Table~\ref{data_quality}, the raters found over 99\% of the sampled  utterances to be natural as well as semantically equivalent to the source English utterance. 

\textbf{\textit{How good are the synthetic CS utterances  when judged by a human?}}
We also performed the same quality measurements on the mT5 generated data to estimate naturalness and semantic equivalence. We randomly sampled 500 generated queries from the small and XXL model trained using the largest seed set. We observed that the data generated by the XXL model was of much higher quality compared to the small model.  For the XXL model, the  raters found over 98\% of the sampled generated utterances to be natural and over 89\% were semantically equivalent to the source English utterance. Additionally, we observed high $\kappa$ ($>$0.65) values for inter-annotator agreement; details  can be found in Table~\ref{data_quality}.

\section{Experimental Setup}

Additionally for extrinsic evaluation, we answer the following research questions through our experiments: \textbf{(a)} What effect does adding augmented data to the training set have on the semantic parsing task? \textbf{(b)} What is the effect of varying the size of initial seed set for the data augmentation step on the overall performance?; \textbf{(c)} How does the performance vary across domains?

To determine the quality of the augmented data, we trained semantic parsing models for code-switched utterances. We fine tuned a second mT5 model for this task. 
 We compared the performance of our semantic parsing models with and without augmented data while keeping everything else the same. To study the efficacy of our technique with regards to the seed set, we trained models by varying the seed set size 100, 500, 1000, 2000 and 3000 utterances. Further, these training data of varying sizes were used to augment data using the un-annotated utterances from the TOPv2 dataset in accordance to section \ref{augment_sec}. Although we only release the augmented data from the model trained on full 3000 training examples, an independent augmented data set was created for each batch of training data from the training split.

For the mT5 models, we use the public checkpoints provided by ~\citet{xue2020mt5}. We only roughly tuned the hparams. For the XXL (13 billion parameters) mT5 model training, we use a learning rate of 0.001, batch size of 512 and  fine tune for 20k steps. For the small mT5 model (300 million parameters), we fine tune for 200k steps keeping rest of the parameters exactly the same. We checked for overfitting by choosing earlier (10k, 50k steps) checkpoints but obtained best results with the parameters mentioned.  

\begin{table}
\footnotesize
\centering
\begin{tabular}{L{3.5cm}L{3.5cm}}
\toprule
\multicolumn{2}{c}{\textbf{Human Annotated Utterances}} \\
\midrule
\textbf{EN:} {Will it be [ \hlcyan{above} ]  [ \hlred{59} ] in [ \hlpurple{celsius} ] at [ \hlorange{Menlo Park} ] [ \hllime{tomorrow morning} ] } & \textbf{CS:} {Kya [ \hllime{kal subah} ] [ \hlorange{Menlo Park} ] me [ \hlred{59} ] [ \hlpurple{celsius} ] se [ \hlcyan{upar} ] hoga?} \\
\midrule
\textbf{EN:} {Remind [ \hlcyan{me} ] [ \hlpurple{on September 1} ] that [ \hlorange{its my best friend's birthday} ] !} &  \textbf{CS:} {[ \hlpurple{September 1 ko} ] [ \hlcyan{muje} ] remind karaye ki [ \hlorange{mere best friend ka birthday he} ] } \\
\midrule
\multicolumn{2}{c}{\textbf{Model Generated Utterances}} \\
\midrule
\textbf{EN:} {I need to change [ \hlcyan{my} ] [ \hlpurple{car maintenance} ] reminder to [ \hlorange{Tuesday 9 AM} ] } &
\textbf{CS:} {Mujhe [ \hlcyan{mere} ] [ \hlpurple{car maintenance} ] reminder ko [ \hlorange{Tuesday subah 9 bajhe ke liye} ] badalne ki zaroorat hai} \\
\midrule
\textbf{EN:} {Text [ \hlcyan{my} ] [ \hlpurple{mom} ] that [ \hlorange{I will um reach home in ten minutes} ] } &
\textbf{CS:} {[ \hlcyan{Meri} ] [ \hlpurple{ma} ] ko text karo ki [ \hlorange{mai um ten minutes me pahuch jaunga} ] } \\
\bottomrule
\end{tabular}
\caption{\label{csparse_examples}
Example of Human annotated CS utterances and Model Generated CS utterances, along with the original English utterance. Same slots have been marked with the same colour across utterances.}
\end{table}

\paragraph{Models used for comparison}

For varying the size of the seed set, we trained three mT5 models. 
\begin{enumerate}[noitemsep,topsep=0pt,parsep=0pt,partopsep=0pt]
    \item A CS utterance generation model, used to generate the augmented data for each seed set. 
    \item A semantic parser trained solely on the seed set.
    \item A semantic parser trained on the seed set and the augmented data.
\end{enumerate}
The models were tested over the test data set from the data annotation task of 6.5k utterances.\footnote{Note that we report all results on this code switched human annotated testset.} We used the exact-match accuracy to judge the performance of these models.

\section{Results} \label{results}
\begin{figure}
    \centering
    \definecolor{alarm}{HTML}{0FA3B1}
\definecolor{timer}{HTML}{ED6A5A}
\definecolor{reminder}{HTML}{685762}
\definecolor{event}{HTML}{E27475}
\definecolor{navigation}{HTML}{000000}
\definecolor{messaging}{HTML}{B38701}
\definecolor{music}{HTML}{4D80BC}
\definecolor{weather}{HTML}{4D80BC}

  \begin{tikzpicture}[scale=0.9]
    \begin{axis}[
      xlabel=\footnotesize \# Human Annotated Examples in Seed Set,
      y label style={at={(axis description cs:0.05,.5)}},
      ylabel=\footnotesize Exact Match Acc.,
      ytick={20,40,60,80,90,100},
      xtick={100,500,1000,2000,3000},
      minor y tick num=9,
      ymax=85,
      ymin=5,
      xmin=100,
      xmax=3000,
      legend style={
        at={(0.5,1.35)},       
        anchor=north,
        legend columns=1,       
        /tikz/every even column/.append style={column sep=0.01cm}
      },
      ]
    \addplot[mark=square,ultra thick,color=alarm] table [x=size, y=baseline, col sep=comma] {ema_data.txt};
    \addlegendentry{Small no augmentation}
    
    \addplot[mark=diamond,ultra thick,color=alarm] table [x=size, y=cst5, col sep=comma] {ema_data.txt};
    \addlegendentry{Small with \name~augmentation}
        
    \addplot[mark=o,very thick,color=event] table [x=size, y=baseline-xxl, col sep=comma] {ema_data.txt};
    \addlegendentry{XXL no augmentation}

    \addplot[mark=triangle,very thick,color=event] table [x=size, y=cst5-xxl, col sep=comma] {ema_data.txt};  
    \addlegendentry{XXL with \name~augmentation}
    \addplot [mark=none, color=black, thick,dotted] coordinates {(100, 60) (2000, 60)};
    \addplot [mark=none,color=black, thick,dotted] coordinates {(2000, 60) (2000, 0)};
    \node[label={270:{20x}},inner sep=2pt] at (axis cs:2150,20) {};
  \end{axis}
\end{tikzpicture}
    \caption{Comparison of Exact Match Accuracy for semantic parsing models trained with and without \name~ augmentation, for varying seed set size. The dotted line shows the XXL model reaches EM of 60\% using \name~ augmentation with 100 examples, equivalent to fine tuning a XXL model with 2k examples. }
    \label{ema}
\end{figure}
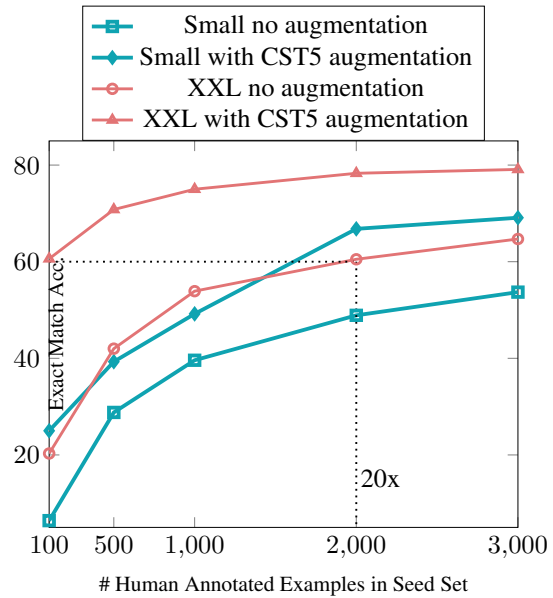
\begin{figure}
\definecolor{alarm}{HTML}{0FA3B1}
\definecolor{timer}{HTML}{ED6A5A}
\definecolor{reminder}{HTML}{685762}
\definecolor{event}{HTML}{E27475}
\definecolor{navigation}{HTML}{000000}
\definecolor{messaging}{HTML}{B38701}
\definecolor{music}{HTML}{4D80BC}
\definecolor{weather}{HTML}{4D80BC}

  \begin{tikzpicture}[scale=0.9]
    \begin{axis}[
      xlabel=\footnotesize \# Training Examples,
      y label style={at={(axis description cs:0.05,.5)}},
      ylabel=\footnotesize Exact Match Acc.,
      ytick={20,40,60,80,90,100},
      xtick={100,500,1000,2000,3000},
      minor y tick num=9,
      ymax=100,
      ymin=20,
      xmin=100,
      xmax=3000,
      legend style={
        at={(0.5,1.3)},       
        anchor=north,
        legend columns=3,       
        /tikz/every even column/.append style={column sep=0.01cm}
      },
      ]
      
    \addplot[mark=o,very thick,color=alarm] table [x=size, y=alarm, col sep=comma] {domain_data.txt};
    \addlegendentry{alarm}
    \addplot[mark=triangle,very thick,color=event] table [x=size, y=event, col sep=comma] {domain_data.txt};
    \addlegendentry{event}
    \addplot[mark=None,very thick,dashed,color=messaging] table [x=size, y=messaging, col sep=comma] {domain_data.txt};
    \addlegendentry{messaging}
    \addplot[mark=None,very thick,loosely dashdotted,color=music] table [x=size, y=music, col sep=comma] {domain_data.txt};
    \addlegendentry{music}
    \addplot[mark=None,very thick,dotted,color=navigation] table [x=size, y=navigation, col sep=comma] {domain_data.txt};
    \addlegendentry{navigation}
    \addplot[mark=None,very thick,densely dotted,color=reminder] table [x=size, y=reminder, col sep=comma] {domain_data.txt};
    \addlegendentry{reminder}
    \addplot[mark=None,very thick,densely dashed,color=timer] table [x=size, y=timer, col sep=comma] {domain_data.txt};
    \addlegendentry{timer}
    \addplot[mark=None,very thick,densely dashdotted,color=weather] table [x=size, y=weather, col sep=comma] {domain_data.txt};
    \addlegendentry{weather}
  \end{axis}
\end{tikzpicture}
\caption{Exact Match Accuracy for different domains in the Hinglish-TOP dataset, with varying amount of seed training examples used for data augmentation.}
\label{domain_plot}
\end{figure}
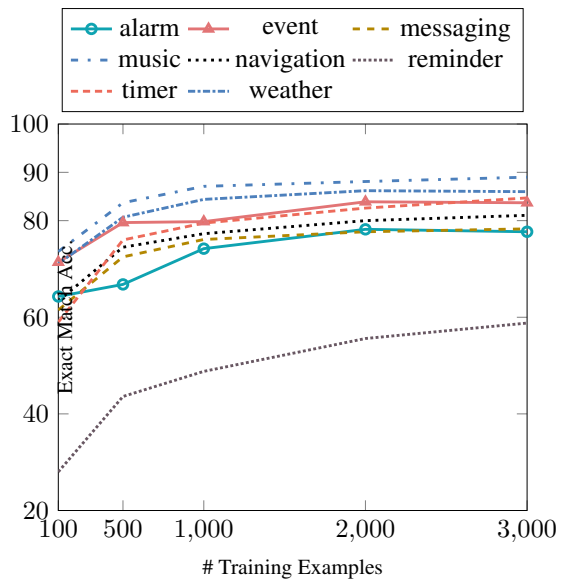

As seen from Figure~\ref{ema}, we observe that increasing the size of the training data, improves the performance of both models(with and without augmented data set) significantly. While there is a consistent improvement on adding more training data, the improvement itself seems to diminish with more seed set examples. Empirically, we observe that the models which make use of the augmented data reaches saturation, i.e. the value after which additional data points don't add to the overall performance of the model, quicker than the models which do not make use of the augmented data. This shows that the synthetic data generated by our technique is useful for the end task of semantic parsing. 

As seen from Figure~\ref{ema}, the performance of the mT5 model trained using the augmented data was consistently better than that trained without it. The XXL model trained using our technique with a seed set of only 100 examples performs similar to the model trained with 2000 human annotated examples. This is over a 40\% absolute improvement with 100 examples. Even with 3000 human annotated examples we see over 15\% improvement in EM accuracy.  This goes to show the effectiveness our data augmentation technique, especially with low-resource settings.

The XXL mT5 model was able to produce a significantly higher number  of  code-switched utterances compared to the small mT5 model, which ultimately led to the significant gain in its performance. The number of generated code-switched utterances post data-filtering can be seen in Table \ref{synthetic_data_stat}. As seen from the results (Figure~\ref{ema}) the XXL mT5 model is significantly better than the small mT5 model, both as a parser and a data generator.   We also observed that adding the augmented data, allows the model to handle more complex queries. A few examples from the models trained on full 3000 training examples have been included in Table~\ref{csparse_examples}.

Moreover, after an in-depth domain-wise analysis of the models performance (Figure \ref{domain_plot}), we observed that the model struggled with the \textit{reminder} domain. It is interesting to note that the \textit{reminder} domain had the highest number of unique intents, 16. Also, reminder domain had the highest number of average intents per utterance of ~2.66. This collectively led to the low performance of the model on this domain. Some of the domains saturated quicker and did not see much improvement upon adding more data points. We can empirically observe that for domains with more inherent difficulty~(reminder for example) the value of added data is is more than simpler domains ~(weather for example).  

\paragraph{Results on CSTOP}Additionally, we repeated a similar experiment for CSTOP~\cite{einolghozati2021volumen}, which contains spanish-english code switched utterances annotated in the TOP schema. Unlike our approach CSTOP doesn't have parallel data with TOPv2. Since only the weather domain was common between CSTOP and TOPV2, we work with the weather domain subset. As we need parallel english data to finetune the mT5 model, we used google translate to generate parallel english data for 100 spanish-english queries. We used this to finetune a mT5 model and generated new synthetic data using TOPv2 weather domain as the monolingual dataset. For semantic parsing using 100 queries from the training data we observe similar trends as Hinglish.  We observe an improvement in the parsing performance from 69.2\% EM to 77.8 \% EM on the testset using the synthetic code switched data . 

\section{Conclusion}
We proposed \name, an approach to overcome the scarcity of labeled data for code-switched semantic parsing. \name~ generates code switched utterances from English utterances using a large pretrained LM mT5, and a small number of seed examples. We showed that the generated utterances were of high quality, as determined by human raters in a quality annotation task. By aligning the generated utterances with the semantic parse of the original English utterance, we derived a large supervised dataset for training a code-switched semantic parser using a labeled dataset in English.

We did both intrinsic and extrinsic evaluation of the human annotated as well as the synthetic data. Human raters found 98\% of the synthetic data to  be natural and 89\% to be semantically equivalent. Extrinsically, our experiments also demonstrated that data augmentation using \name~ effectively reduces the data requirements by 20x for the code switched semantic parsing task. To further research in this area, we will release the dataset of over 10k manually annotated Hinglish utterances, along with over 170k examples generated using \name\footnote{This data is available at \url{https://github.com/google-research-datasets/Hinglish-TOP-Dataset}}.  While we applied \name~  to Hindi-English code-switching, our approach is general and can be applied to any other language pair that exhibits code-switching. In the spirit of \name, we believe that using large pre-trained language models to perform data augmentation for other code-switched NLP tasks is an attractive future direction to explore.

\bibliography{custom}
\bibliographystyle{acl_natbib}

\end{document}